\documentclass{article}

\usepackage[table]{xcolor} 
\usepackage[preprint, nonatbib]{neurips_2024}
\usepackage{amsmath}
\usepackage[utf8]{inputenc}
\usepackage{algorithm}
\usepackage{algpseudocode}
\usepackage{comment}
\usepackage{wrapfig}
\usepackage{tablefootnote}
\usepackage{caption}
\usepackage{subfigure}
\usepackage{pythonhighlight}
\usepackage[export]{adjustbox}

\usepackage[utf8]{inputenc} 
\usepackage[T1]{fontenc}    
\usepackage[hidelinks]{hyperref}        
\usepackage{url}            
\usepackage{booktabs}       
\usepackage{tabularx} 
\usepackage{amsfonts}       
\usepackage{nicefrac}       
\usepackage{microtype}      
\usepackage{xcolor}         
\usepackage{multirow}
\usepackage{enumitem}
\usepackage{epstopdf}
\usepackage{graphicx}
\usepackage{array}    

\title{HQ-DiT: Efficient Diffusion Transformer with FP4 Hybrid Quantization}

\author{
  Wenxuan Liu \\
  New York University \\
  \texttt{wl3181@nyu.edu} \\
  \And
  Sai Qian Zhang \\
  New York University \\
  \texttt{sai.zhang@nyu.edu} \\
}

\begin{document}
\maketitle

\begin{abstract}
    Diffusion Transformers (DiTs) have recently gained substantial attention in both industrial and academic fields for their superior visual generation capabilities, outperforming traditional diffusion models that use U-Net.
    However, the enhanced performance of DiTs also comes with high parameter counts and implementation costs, seriously restricting their use on resource-limited devices such as mobile phones. To address these challenges, we introduce the~\textit{Hybrid Floating-point Quantization for DiT} (HQ-DiT), an efficient post-training quantization method that utilizes 4-bit floating-point (FP) precision on both weights and activations for DiT inference. Compared to fixed-point quantization (e.g., INT8), FP quantization, complemented by our proposed clipping range selection mechanism, naturally aligns with the data distribution within DiT, resulting in a minimal quantization error. Furthermore, HQ-DiT also implements a universal identity mathematical transform to mitigate the serious quantization error caused by the outliers. The experimental results demonstrate that DiT can achieve extremely low-precision quantization (i.e., 4 bits) with negligible impact on performance. Our approach marks the first instance where both weights and activations in DiTs are quantized to just 4 bits, with only a 0.12 increase in sFID on ImageNet $256\times 256$. 
\end{abstract}

\section{Introduction}
\label{sec:intro}
Diffusion Transformers (DiTs)~\cite{dit} have garnered increasing attention due to their superior performance over traditional diffusion models (DMs) that use U-Net~\cite{ldm} as the backbone DNN. Since their introduction, they have been extensively researched and applied in both academic and industrial fields~\cite{dit, mo2024dit, feng2024latent, wu2024medsegdiff, gao2023masked, chen2023pixart}, with the most notable application being OpenAI's SoRA~\cite{Sora}. Recent research has demonstrated its impressive generative capabilities across various modalities~\cite{gao2024lumina}. 
However, the iterative denoising steps and massive computational demands significantly slow down its execution. Although various methods~\cite{dm_method2, dm_method1} have been proposed to reduce the thousands of iterative steps to just a few dozen, the large number of parameters and the complex network structure of DiT models still impose a significant computational burden at each denoising timestep. This hinders their applicability in practical scenarios with limited resource constraints.

\begin{wrapfigure}{r}{0.45\textwidth}
    \centering
    \includegraphics[width=0.45\textwidth]{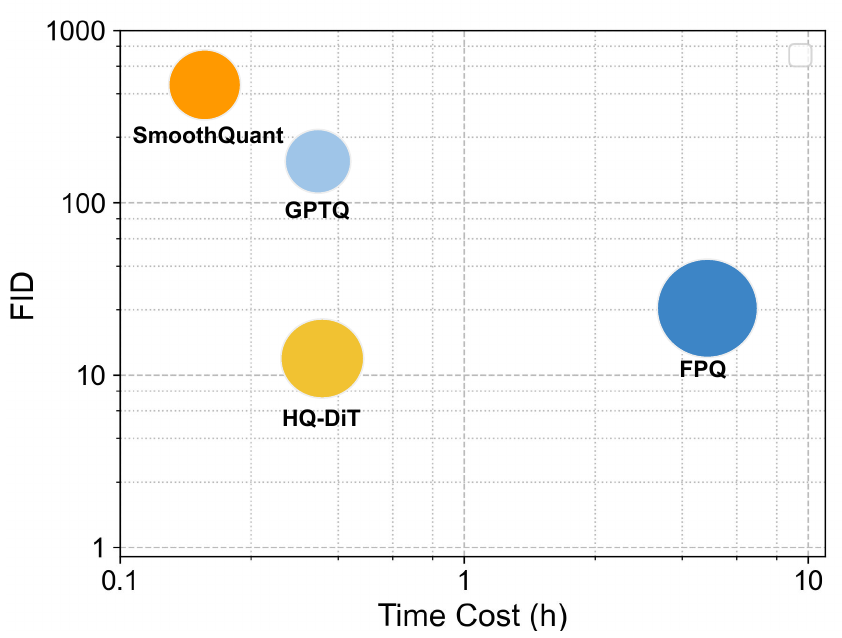}
\caption{Performance of different approaches on ImageNet $256 \times 256$. Both weights and activations are quantized with 4 bits. The x-axis denotes the runtime for each quantization approach. The size of the circle indicates the standard deviation.}
    \label{fig:motivation}
\end{wrapfigure}

Model quantization is widely recognized as an effective approach for reducing both memory and computational burdens by compressing weights and activations into lower-bit representations. Among various quantization methods, Post-training quantization (PTQ) offers a training-free approach (or minimal training cost for calibration purposes~\cite{nagel2020ptq, li2021brecq, lin2021fq, kung2020term, kung2020term1}) for fast and effective quantization. Compared to Quantization-Aware Training (QAT), which requires multiple rounds of fine-tuning, PTQ incurs significantly lower computational costs. This makes it an appealing solution for quantizing large models like DiT. Existing PTQ methods for DMs~\cite{qdiffusion, efficientdm} primarily employs fixed-point quantization (i.e., INT quantization); however, significant quantization errors can occur at low precision. To demonstrate this, we evaluate several recent quantization methods for large models, including SmoothQuant~\cite{smoothquant}, FPQ~\cite{zhang2023afpq}, and GPTQ~\cite{frantar2023gptq} on DiT. As shown in Figure~\ref{fig:motivation}, quantizing both weights and activations to 4-bit precision results in serious performance degradation, leading to an observed increase in the FID up to 100. Moreover, the quantization process involved in these methods also incurs high computational costs due to the calibration process required to search for the optimal quantization scheme.


Floating-point (FP) quantization presents a more flexible alternative to INT quantization. Compared to the integer numeric format, which uses a fixed scaling factor, FP quantization is adaptive to different scales in the data due to the inclusion of an exponent. This adaptability allows it to maintain precision across different magnitudes~\cite{adaptivefp, zhang2022fast}, making FP an ideal data format choice for various commercial hardware platforms, such as Nvidia's Blackwell (H200)~\cite{H200}. 
Unlike INT quantization, which obtains quantized values through truncation and rounding, the challenge in FP quantization lies in selecting the appropriate composition, specifically how many bits are assigned to the exponent and mantissa. Inappropriate selections can result in suboptimal performance. To address this issue, we propose to determine the FP composition based on the channel-wise data distribution. Compared to previous methods~\cite{llmfp4}, our approach optimizes performance by tailoring to the unique characteristics of the data in each channel.

Due to their high inter-channel variance and low intra-channel variance, quantizing the activations is well known to be challenging~\cite{dettmers2022llmint8}. 
Previous works, such as SmoothQuant~\cite{smoothquant}, aim to migrate the difficulty the quantization of activation to weight.
However, these methods still suffer from performance degradation at lower precision levels. In this study, we push the limits of quantization bitwidth by using random Hadamard transforms to eliminate outliers in the activations. Additionally, we apply a corresponding transform to the network weights to minimize quantization errors across the network. This approach effectively reduces the impact of outliers on quantization while introducing only a minimal increase in computational overhead.
To this end, we introduce a hybrid FP Quantization for DiT (HQ-DiT), a simple yet efficient PTQ method designed for low-precision DiT. HQ-DiT achieves performance levels comparable to full-precision models while operating with precision as low as FP4.
Our contributions are summarized as follows:
\begin{itemize}[itemsep=0.5em, topsep=0.5em, parsep=0.5em, leftmargin=*]
\item We introduce HQ-DiT, an efficient PTQ method for DiTs, capable of achieving performance on par with full-precision models using 4-bit FP (FP4) quantization. To the best of our knowledge, HQ-DiT represents the first attempt to quantize the DiT using FP data format. 
\item We propose a novel algorithm that can adaptively select the optimal FP format based on the data distribution, effectively addressing the significant computational overhead associated with search-based methods in the prior work~\cite{zhang2023afpq}.
\item HQ-DiT quantifies both weights and activations in DiTs using FP4, leading to a $5.09\times$ speedup and $2.13\times$ memory savings compared to the full-precision model. Our HQ-DiT achieves state-of-the-art results in low-precision quantization, with the FP4 model outperforming the full-precision latent diffusion model (LDM) in both Inception Score (IS) and Frechet Inception Distance (FID).
\end{itemize}

\section{Background and Related Work}
\label{sec:bg_related_work}

\subsection{Diffusion Models}
\label{sec:bg:dm}
Diffusion models (DMs)~\cite{ho2020denoising} have recently gained significant attention for its remarkable ability to generate diverse photorealistic images. It is a parameterized Markov chain trained through variational inference to generate samples that match the data distribution over a finite duration. Specifically, during the~\textit{forward process} of DMs, given an input image $x_{0}\sim q(x)$, a series of Gaussian noise is generated and added to the $x_{0}$, resulting in a sequence of noisy samples $\{x_{t}\}, 0\leq t \leq T$.
\begin{equation}
    q(x_{t}|x_{t-1}) = \mathcal{N}(x_{t};\sqrt{1-\beta_{t}}x_{t-1}, \beta_{t} I)
\end{equation}
where $\beta_{t}\in (0,1)$ is the variance schedule that controls the strength of the Gaussian noise in each step.

During the~\textit{reverse process}, given a randomly sampled Gaussian noise $\mathcal{N}(x_{T}; 0,I)$. The synthetic images are generated progressively with the following procedure:
\begin{equation}
    p_{\theta}(x_{t-1}|x_{t}) = \mathcal{N}(x_{t-1}; \mu_{\theta}(x_{t},t), \hat{\beta_{t}} I)
\end{equation}
where $\mu_{\theta}(x_{t},t)$ and $\hat{\beta_{t}}$ are defined as follows:
\begin{equation}
    \label{eqn:mean}
    \mu_{\theta}(x_{t},t) = \frac{1}{\sqrt{\alpha_{t}}}(x_{t}-\frac{1-\alpha_{t}}{\sqrt{1-\bar{\alpha}_{t}}}\epsilon_{\theta, t}),\; 
    \hat{\beta_{t}} = \frac{1-\bar{\alpha}_{t-1}}{1-\bar{\alpha}_{t}}
\end{equation}

\begin{wrapfigure}{r}{0.28\textwidth}
    \centering
    \includegraphics[width=0.28\textwidth]{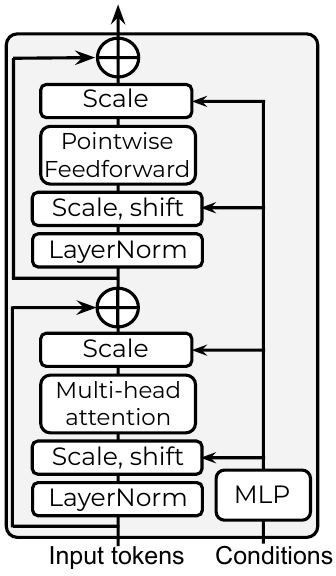}
\caption{A DiT block.}
    \label{fig:dit_block}
\end{wrapfigure}

In equation~\ref{eqn:mean}, ${\alpha}_{t}=1-\beta_{t}$, and $\bar{\alpha}_{t}=\prod_{i=1}^{t} {\alpha}_{t}$. $\epsilon_{\theta, t}$ denotes the predicted noise that is generated with a backbone DNN. The backbone DNN in the conventional DMs utilize a convolutional-based U-Net architecture~\cite{ldm, adm}. In contrast, the DiT model substitutes this U-Net with a transformer and incorporates a parallel MLP that generates scale and bias for intermediate outputs, as shown in Figure~\ref{fig:dit_block}. DiT is recognized for surpassing conventional DM in terms of visual generation capability.

\subsection{Classifier-free Guidance}
\label{sec:bg:cfg}
In DiT, an additional class label $c$ can be provided by the user as guidance for image generation. In this situation, the reverse process becomes:
\[
p_\theta(x_{t-1}|x_t, c)
\]
Classifier-Free Guidance~\cite{ho2022classifier} uses an implicit classifier to replace an explicit classifier, adjusting the guidance weight to control the realism and balanced diversity of generated images. According to Bayes' formula, the gradient of the classifier can be formulated as:
\[
\nabla_{x_t} \log p(c \mid x_t) = \nabla_{x_t} \log p(x_t \mid c) - \nabla_{x_t} \log p(x_t)
\]
\[
= \frac{-1}{\sqrt{1-\bar{\alpha}_t}} \left( \varepsilon_\theta (x_t, t, c) - \varepsilon_\theta (x_t, t) \right)
\]
By interpreting the output of DMs as the score function, the DDPM sampling procedure~\cite{ho2020denoising, sohldickstein2015deep} can be guided to sample $x$ with high probability $p(x \mid c)$ by:
\[
\bar{\varepsilon_\theta}(x_t, c) = \varepsilon_\theta(x_t, \emptyset) + s \cdot (\varepsilon_\theta(x_t, c) - \varepsilon_\theta(x_t, \emptyset)) + s \cdot \nabla_{x} \log p(c \mid x_t) \propto \theta(x_t, \emptyset)
\]

where $s > 1$ indicates the scale of the guidance. Classifier guidance can control the balance of realism and diversity of generated samples and is widely applied in the generative models such as DALL$\cdot$ E~\cite{ho2022classifier, nichol2021glide, ramesh2022hierarchical}.

\begin{figure*}[t]
    \centering
    \includegraphics[width=\linewidth]{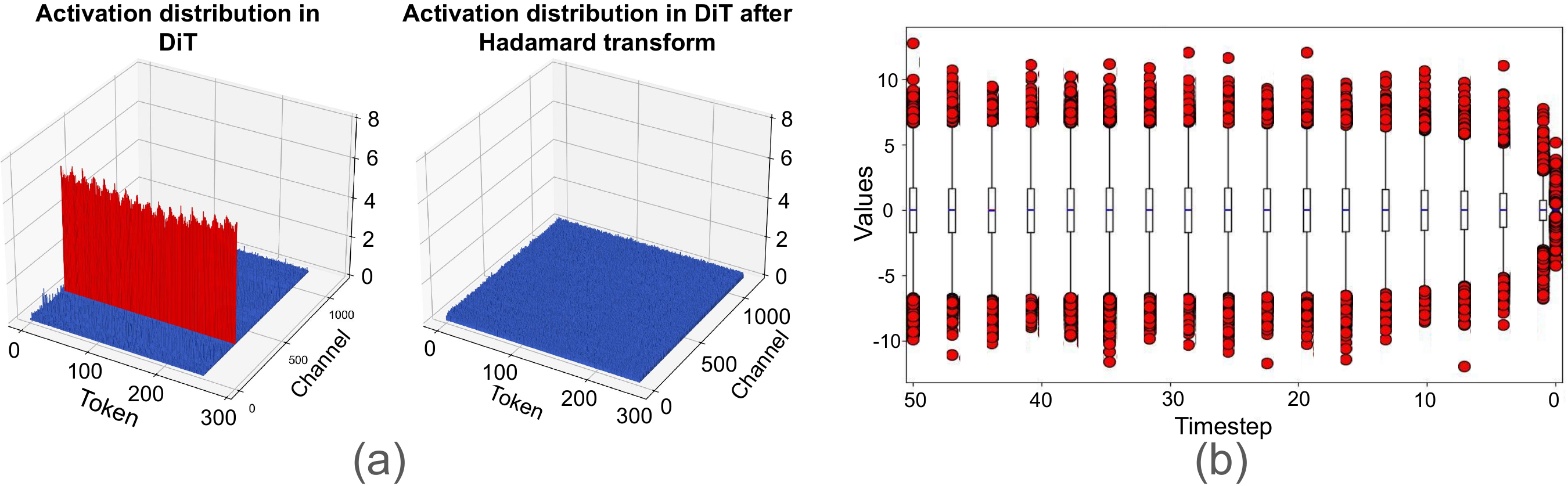}
    \caption{(a) Magnitude distribution of an input activation of a DiT linear layer before and after Hadamard transform. (b) Histogram on an input activation matrix across different time steps.}
    \label{fig:activation}
\end{figure*}

\subsection{Post-Training Quantization}
\label{sec:bg:ptq}
Unlike QAT, PTQ does not require model training or incurs only minimal training costs for calibration purposes, making it very efficient in computation. Furthermore, when calibrating generative models such as DMs, instead of using the original training dataset, calibration datasets can be generated using the full-precision model. This enables the calibration process to be implemented in a data-free manner. For example, Q-Diffusion~\cite{qdiffusion} applies advanced PTQ techniques proposed by BRECQ~\cite{li2021brecq} to improve performance and evaluate it on a wider range
of datasets, while PTQD~\cite{ptqd} further breaks down quantization error and combines it into diffusion noise. EfficientDM~\cite{efficientdm} enhances the performance of the quantized DM by fine-tuning the model using QALoRA~\cite{xu2023qa, han2024parameter}. However, all of the aforementioned methods primarily focus on INT quantization in conventional DMs with U-Net architectures, while HQ-DiT aims for quantizing DiT with low-precision FP numeric format. FPQ~\cite{llmfp4} determines the optimal FP bias and format by exhaustively search over the design space, leading to a FP4 LLM with significant computational cost on quantization process. In contrast, our method calculates the optimal format based on data distribution and employs random Hadamard transforms to mitigate outliers, achieving superior performance with negligible cost on the quantization process.

FP quantization has been widely utilized as an alternative to INT quantization for improving the efficiency of deep neural networks (DNNs). Compared with INT quantization, FP data format has a wider dynamic range, making them better adapt to the wide-spread data distribution in DNNs. Multiple customized FP formats have been proposed and adopted in commercial hardware, including Google's bfloat16~\cite{bfloat} and Nvidia's TensorFloat 32~\cite{tf32}. Recently, Nvidia announced that FP4 and FP8 will also be supported in the new series of Blackwell GPUs. FP formats have also shown to be efficient in quantizing LLM~\cite{llmfp4, zhang2023afpq}. In this work, we aim for applying FP for quantizing DiT.

\section{Methodology}
\label{sec:methodology}
In this section, we describe the HQ-DiT method in detail. We first study the data distribution within DiT in Section~\ref{sec:input-distribution}. Next we describe Hadamard transform and its application in DiT in Section~\ref{sec:act-quant}, followed by the detailed quantization strategies in Section~\ref{sec:weight-quant} and Section~\ref{sec:overall-quant}.

\subsection{How Activations are Distributed within DiT?}
\label{sec:input-distribution}
\begin{wrapfigure}{L}{0.52\textwidth}
\vspace{-2.5em}
\begin{minipage}{0.52\textwidth}
\begin{algorithm}[H]
\caption{MinMax quantization}
\begin{algorithmic}
\State \textbf{Input:} FP32 array $A_{fp}$, number of bits $n$, number of exponent bits $n_e$.
\State $n_m \gets n - n_e - 1$ \Comment{Calculate mantissa bitwidth}
\State $max\_val \gets 2^{(2^{n_e} - 1)} \times (2 - 2^{-n_m})$. 
\State $A_{sign} \gets \text{sign}(A_{fp})$.
\State $A_{abs} \gets \text{abs}(A_{fp})$.
\State $bias \gets \left\lfloor \log_2(\max(A_{abs})) - \log_2(max\_val) \right\rfloor$
\State $value\_max \gets 2^{(2^{n_e} + bias - 1)} \times (2 - 2^{-n_m})$
\State $A_{abs} \gets max(A_{abs}, value_{max})$
\State $x_{log\_scales} \gets \text{clamp}(\left\lfloor(\log_2(A_{abs}) - bias\right) \rfloor, 1)$
\State $scale \gets 2^{(x_{log\_scales} - n_m + bias)}$

\State $A'_{abs} \gets \text{quantize and round } A_{abs} \text{ by } scale$ 
\State $A'_{fp} \gets A'_{abs} * A_{sign}$ \Comment{Apply quantization}
\State \Return $A'_{fp}$ \Comment{Return the quantized array}
\end{algorithmic}
\label{alg:fp_quantization}
\end{algorithm}
\end{minipage}
\end{wrapfigure}
To understand the distribution of the input activation within DiT, we collect the input activations of a DiT block across 50 denoising steps, the histogram is highlighted in Figure~\ref{fig:activation} (b). We then analyze the distribution of the activation matrix at a specific time step, as shown in the left part of Figure~\ref{fig:activation} (a). Similar to the pattern observed in DMs~\cite{qdiffusion}, outliers exist at the level of entire channels. Specifically, the scale of outliers in activations is approximately $100 \times$ larger than rest of the activation values. Thus, per-token quantization would inevitably introduce substantial errors. Given that activations exhibit high variance across different channels but low variance within channels, with outliers confined to a limited number of channels, implementing per-channel quantization and minimizing the impact of outliers can hopefully result in lower quantization error.

\subsection{Hadamard Transform for Activation Quantization}
\label{sec:act-quant}
Previous work~\cite{quip} has demonstrated that Hadamard transform can be applied to eliminate outliers present in data. Similarly, we introduce random Hadamard transforms to eliminate outliers present in the input activations $X$ of DiT by multiplying it with an orthogonal Hadamard matrix $H \in \mathbb{R}^{n \times n}$, where $n$ is the embedding dimension of DiT and $H$ satisfies $HH^{\top}= H^{\top}H = I$. The resultant matrix $XH$ will exhibit a much smoother distribution, with most outliers being eliminated, as depicted in the right part of Figure~\ref{fig:activation} (a). To maintain the mathematical equivalence of the linear layer within the DiT block, we need to apply the Hadamard matrix to the corresponding weight matrices within the self-attention (SA) layers and pointwise feedforward (FFN) layers of each DiT block. Next we will describe these modifications in detail.

\begin{figure}[t]
    \centering
    \includegraphics[width=\linewidth]{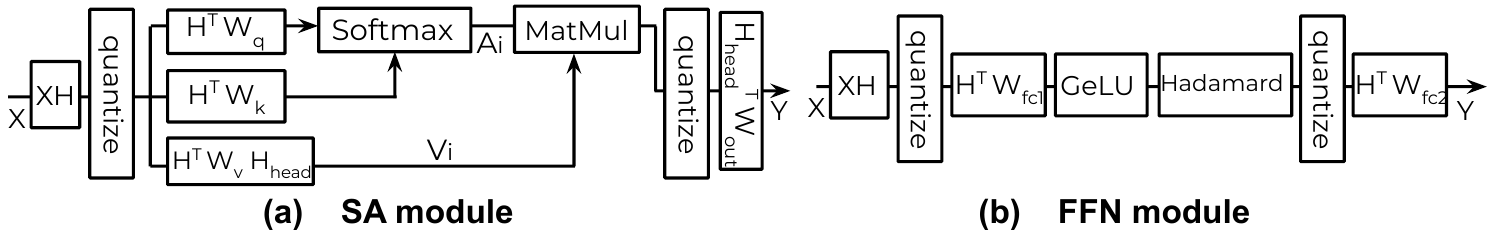}
    \caption{Quantization workflow in (a) SA block and (b) FFN block for DiT.}
    \label{fig:attn_block}
\end{figure}

\paragraph{Hadamard Transform within Self-attention Module} To achieve computational invariance, each of the query, key, and value weight matrices $W_q$, $W_k$, and $W_v$ within the SA block are multiplied by $H^{\top}$, producing new matrices $W'_q = H^\top W_{q}$, $W'_{k}= H^\top W_{k}$, and $W'_{v}= H^\top W_{v}$, which can be performed prior to the DiT execution and incurs no additional online computational cost. This ensures that the Hadamard transform applied to the input activations will not alter the output, as $(XH)(W') = (XH)(H^\top W)=XW$. This will produce queries, keys, and values that are mathematically equivalent to the original outputs. The resultant input activations $XH$ can then be quantized with much smaller quantization error. This process is highlighted in Figure~\ref{fig:attn_block} (a).

After that, the attention matrix of each head $A_{i}$, which is the output of the softmax layer, is multiplied by the corresponding value matrix $V_{i}$. The results are then multiplied by the $W_{out}$ matrix, as shown in Figure~\ref{fig:attn_block} (a). This process produces the SA block output $Y =  VW_{out}$, where $V=\text{concat}[A_{1}V_{1}, \ldots, A_{h}V_{h}]$ and $h$ denotes the number of heads. However, outliers will also be present within $V$, necessitating the application of the Hadamard transform to $V$. 
In order to apply Hadamard transform over $V$ without any additional multiplication computation, we need to fuse the Hadamard matrix $H_d$ on each head, where $H_d$ denotes a $d\times d$ Hadamard matrix, and $d$ is the embedding size of each head. Since each component of $V$ are first produced at each head and then concatenated together to form $V$, which are then multiplied against $W_{out}$, we can apply Hadamard transform on $W_v$ and $W_{out}$ with $H_{head} $ by leveraging the properties of the Kronecker product.
\[
W_v \leftarrow W_v H_{head}, \quad W_{\text{out}} \leftarrow H_{head}W_{\text{out}},
\]
where $H_{head} = ((I \otimes H_{d})(H_{h} \otimes I)) $, $I$ is the identity matrix, $H_d$ denotes a $d\times d$ Hadamard matrix, $H_h$ denotes a $h\times h$ Hadamard matrix, $h$ is the number of heads. $\otimes$ represents the Kronecker product. This effectively mitigates the outliers within $V$ without any additional computational overhead.

\paragraph{Hadamard Transform within FFN Module} The FFN module within DiT consists of two linear layers and one GELU activation function (Figure~\ref{fig:attn_block} (b)). Denote $W_{fc1}$ and $W_{fc2}$ the weight matrices associated with these two linear layers. To better quantize the intermediate activations, we also fuse the Hadamard transform matrices into $W_{fc1}$ by executing $W_{fc1} \leftarrow H^\top W_{fc1}$. Similarly, this operation can be performed offline without any additional cost during the inference process.
The output of the first linear layer $W_{fc1}$ also contains outliers and an additional round Hadamard transform are required to remove the outliers. Due to the presence of the GELU activation function, the Hadamard transform need to be performed online. To mitigate this cost, we have designed an efficient Hadamard transform with the following computational complexity:

\textbf{Lemma 1} \textit{If the dimension \( n \) of \(X\) is a power of 2, the Walsh-Hadamard transform allows us to compute \( XH \) in \( O(n^2 \log n) \) time complexity. If \( n \) is not a power of 2, by leveraging the properties of the Kronecker product, we can complete the computation in \( O(n^2 q \log p) \) time complexity, where \( p \) is the largest power of 2 such that we can construct a new Hadamard matrix by the Kronecker approach.}

We will provide a detailed description of Lemma 1 in the appendix. Applying the Hadamard transform to activations and weights can effectively eliminate outliers. Next, we discuss the method used for quantizing weights and activations.

\subsection{FP Format Selection for Weight Quantization}
\label{sec:weight-quant}
\begin{figure}[t]
\centering
\includegraphics[scale=0.6]{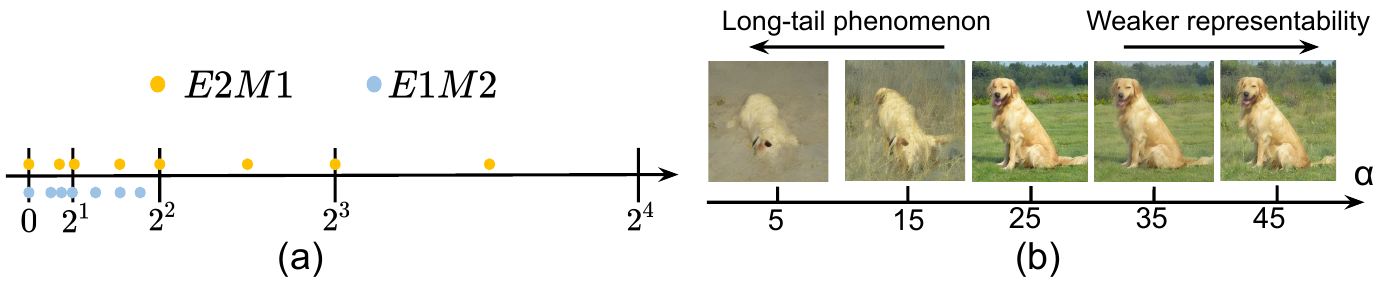}
\caption{(a) Distribution of FP format E2M1 and E1M2, with the bias set to 0. (b) Effect of $\alpha$ on FP format selection. If \(\alpha\) is too small, a larger \(n_e\) will be chosen, leading to severe data distribution imbalances. If \(\alpha\) is too big, less number can be represented. Samples generated on W4A4 ImageNet 256 $\times$ 256 (cfg=1.5).}
\label{figure:alpha}
\end{figure}

In this section, we describe the approach for quantizing the weight matrices within the DiT blocks. Our objective of post-training quantization is to find a quantized weight matrix $\hat{W}$ which minimizes the squared error, relative to the full precision layer output. Formally, this can be formulated as:
\[
\arg\min_{\hat{W}} \|\hat{W}X - WX\|_2^2
\]
\begin{wrapfigure}{r}{0.45\textwidth}
    \centering
    \includegraphics[width=0.45\textwidth]{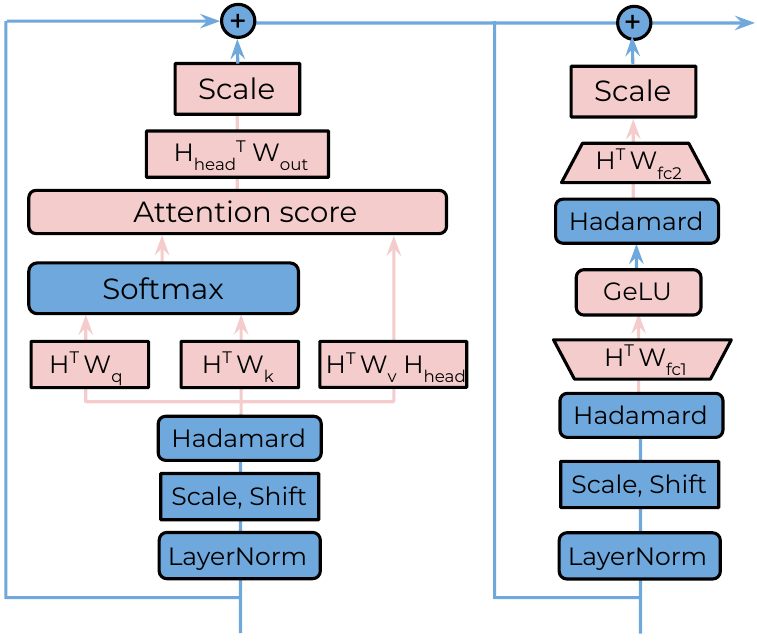}
\caption{HQ-DiT quantization scheme within a DiT block. Operations in FP4 and FP32 are highlighted in blud and red, respectively.}
    \label{fig:workflow}
\end{wrapfigure}
Selecting an appropriate FP composition for weight quantization is crucial, as an improper choice on exponent and mantissa bitwidth can lead to significant quantization errors. A naive approach to selecting an FP data format involves exhaustively searching through each possible combination, which results in high computational overhead. In this work, we propose a simple yet effective method for FP format selection. Our approach is based on the simple fact that for FP format with a fixed total bitwidth, a greater number of exponent bits allows for a larger range of representable values. As shown in Figure~\ref{figure:alpha} (a), a large exponent bitwidth (e.g., E2M1) will also result in an uneven distribution of the represented data, making the long-tail phenomenon more pronounced. For the weight matrices in the DiT model, we can determine the optimal FP format by analyzing their data distribution via the following indicator:
\[
s_{w} = \frac{\max(|W|)}{\text{Quantile}(|W|, \alpha)}
\]
where $\alpha$ is a hyperparameter that return the bottom $\alpha$ percentile of the weight values. By setting $\alpha$ to be a small number (i.e., 10), we can approximate the ratio of the maximum and minimum element within $W$, while eliminating the impact of the outliers. Our goal is to represent all the elements within this interval with low error. On the other hand, for a given FP format with exponent and mantissa bitwidths of $n_e$ and $n_m$, respectively, the ratio between the maximum and minimum values can be computed as follows:
\[
r = 2 \times \frac{max\_val}{min\_val} = 2^{(2^{n_e})} \times \frac{(2 - 2^{-n_m})}{(1 + 2^{-n_m})}
\]
where $max\_val=2^{(2^{n_e} - 1)} \times (2 - 2^{-n_m})$ and $min\_val=(1 + 2^{-n_m})$ are the maximum and minimum values this FP format can represent. To maximize the representation of elements within an interval, we aim to find the optimal FP number format to bring $r$ in close approximation to $s_w$. This is easily achievable due to the limited number of FP format combinations available for a given total bitwidth.

Based on the evaluation results of DiT models, we observe that $\alpha = 25$ yields the best performance. The effect of \(\alpha\) selection over the image quality can be seen in Figure~\ref{figure:alpha} (b). We then apply the FP composition searching methods described earlier to find the optimal FP composition for each DiT block. This results in a hybrid FP data format that significantly enhances performance, as demonstrated by the evaluation results in Section~\ref{sec:ablation}. 
After an appropriate composition is found, we then use the GPTQ~\cite{frantar2023gptq} approach to perform the FP quantization. GPTQ is originally designed for linear quantization, and we modify it to support FP quantization, the implementation details can be found in the appendix.

\subsection{Activation Quantization}
\label{sec:overall-quant}
Given the need to quantize activations in real-time, employing a complex quantization scheme would incur significant computational overhead. Therefore, following the application of the Hadamard transform to the input activation, we utilize the straightforward MinMax quantization method for activation, as described in Algorithm~\ref{alg:fp_quantization}. MinMax quantization sets the bias bit based on the maximum value in each channel of activation, then maps full-precision data to the quantized data range by dividing the scale factor. Evaluation results demonstrate that MinMax quantization can deliver excellent performance on activation quantization with minimal computational overhead.

As depicted in Figure~\ref{fig:workflow}, HQ-DiT is applied to all the fully-connected layers within each DiT block. Computations conducted in low-precision FP representations are highlighted in blue, while the remaining operations, highlighted in red, are performed with FP32.




\section{Experiments}
\label{sec:experiment}
In this section, we evaluate the performance of HQ-DiT in generating both unconditional and conditional images on ImageNet at different resolutions, including $256\times 256$ and $512 \times 512$. We obtain the pretrained DiTs from the official Github repository~\cite{DiT1}, apply the quantization techniques detailed in Section~\ref{sec:methodology}, and evaluate the performance of the quantized DiT.
To compare the performance of HQ-DiT, we adopt the~\textit{Minmax} algorithm described in Algorithm~\ref{alg:fp_quantization} as a baseline, which performs channel-wise FP quantization on both weight and activation. Furthermore, we also implement other sophisticated quantization approaches over DiT including FPQ~\cite{llmfp4}, SmoothQuant~\cite{smoothquant} and GPTQ~\cite{frantar2023gptq}, details can be found in appendix. We evaluate different approaches using the criteria including Inception score (IS), FID and sFID~\cite{ho2022video} by sampling over 50K images, subsequently using ADM's TensorFlow Evaluation suite~\cite{adm} to produce the final results. 

\subsection{Main Results}
\label{sec:main_results}
In this section, we present results on both conditional and unconditional image generation using HQ-DiT. In unconditional image generation~\cite{ho2022classifier}, no external guidance is provided during the iterative DiT denoising process. 
For conditional image generation~\cite{zhang2023adding}, specific class labels are provided as an additional input to guide the DiT denoising process.

\paragraph{Evaluation on unconditional image generation:}
The evaluation results for unconditional image generation are shown in Table~\ref{tab:uncondi_results}. While SmoothQuant and FPQ are effective for 8-bit weight and 8-bit activation (W8A8), their performance deteriorates significantly with 4-bit weight and 4-bit activation. In contrast, HQ-DiT demonstrates superior performance, achieving the highest IS and FID at W8A8 precision. Furthermore, it shows only a slight increase in FID compared to the full-precision (FP32) results. Additionally, our approach maintains the capability to perform unconditional generation tasks even at W4A4 precision, whereas the other two methods struggle to produce images that are recognizable to humans. This significant improvement highlights the notable capability of HQ-DiT to mitigate the impact of outliers and select the optimal data format for each layer. This results in low-precision quantization with minimal quantization error. Additional generation results can be found in the appendix.

\begin{table}[ht]
\centering
\caption{Quantization results for unconditional image generation with input size of $256\times 256$.}
\label{tab:model_performance}
\begin{adjustbox}{width=0.7\columnwidth, center}
\begin{tabular}{@{}ccccccc@{}} 
\toprule
Method & Bit-width (W/A) & IS $\uparrow$ & FID $\downarrow$ & sFID $\downarrow$ \\
\midrule
  FP32        & 32/32  & 112.73 & 12.54 &9.38\\
\midrule
MinMax (FP)   & 8/8  &72.3  & 40.06& 47.23\\
FPQ & 8/8   & 105.76  & 13.52 & 9.62 \\
SmoothQuant (FP)       & 8/8    & 110.17 & 13.23 & \textbf{9.40}\\
\cellcolor{green!25} HQ-DiT      & \cellcolor{green!25}8/8    & \cellcolor{green!25} \textbf{111.62} & \cellcolor{green!25} \textbf{12.74} & \cellcolor{green!25} 9.42 \\
\midrule
MinMax (FP)   & 4/4  & 1.52 & 300.64 & 423.50 \\
FPQ & 4/4  & 8.69 & 231.83 & 168.72 \\
SmoothQuant (FP)       & 4/4   & 4.82 & 289.21 & 204.37 \\
\cellcolor{green!25} HQ-DiT      & \cellcolor{green!25} 4/4     & \cellcolor{green!25} \textbf{61.34} & \cellcolor{green!25} \textbf{23.91} & \cellcolor{green!25} \textbf{17.02} \\
\bottomrule
\end{tabular}
\end{adjustbox}
\label{tab:uncondi_results}
\end{table}

\paragraph{Evaluation on conditional image generation:} 
We also evaluate the performance of HQ-DiT for class-conditional image generation, as detailed in Table~\ref{tab:condi_results}. Experiments on ImageNet $256 \times 256$ are conducted with two different Classifier-Free Guidance Scale (cfg), 1.5 and 4.0. Similarly, we provide comprehensive evaluation results with three metrics including IS, FID and sFID. We observe that for a cfg of 4.0, compared with the FP32 DiT, HQ-DiT achieves an average reduction of 6.98 in FID and 2.22 in sFID for W4A8. HQ-DiT also achieves the optimal performance compared with other baseline algorithms. For W4A4, other methods struggle to mitigate the substantial quantization noise caused by low-bit quantization. For example, FPQ at W4A4 bitwidth experiences a decrease in IS by 248.73 and an increase in sFID by 9.96. In contrast, HQ-DiT can still achieving a high IS of 437.13 and a low sFID of 9.94. Meanwhile, our model gets a 66.44 increase in IS and 0.19 decrease compared with full-precision LDM model~\cite{ldm}.

\begin{table}[ht]
\centering
\caption{Results for conditional image generation on ImageNet $256\times256$ and ImageNet $512\times 512$.}
\begin{adjustbox}{width=0.8\columnwidth,center,}
\begin{tabular}{@{}cccccccccc@{}} 
\toprule
Model & Method & Bit-width (W/A) & IS $\uparrow$ & FID $\downarrow$ & sFID $\downarrow$\\
\midrule
\multirow{9}{*}{\parbox{3cm}{\centering DiT-XL/2 \\$256 \times 256$\\ (steps = 100 \\ cfg = 1.5)}} & FP32 & 32/32 & 266.57 & 2.55~\tablefootnote{The original DiT paper~\cite{dit} reports a FID of 2.27, but our replication efforts yield an FID of 2.55.} & 5.34  \\
                                   & SmoothQuant & 4/8 & 12.84 & 137.60 & 81.22  \\
                                   & SmoothQuant & 4/4 & 6.93 & 252.34 & 192.87  \\
                                   & FPQ & 4/8 & 108.81 & 20.33 & 17.11  \\
                                   & FPQ & 4/4 & 12.62 & 127.99 & 69.10  \\
                                   & GPTQ & 4/8 & 15.09 & 107.63 & 75.33 \\
                                   & GPTQ & 4/4 & 3.14 & 200.85 & 329.18 \\
                                   & \cellcolor{green!25}HQ-DiT & \cellcolor{green!25}4/8 & \cellcolor{green!25}\textbf{145.48} & \cellcolor{green!25}\textbf{12.59} & \cellcolor{green!25}11.41  \\
                                   & \cellcolor{green!25}HQ-DiT & \cellcolor{green!25}4/4 & \cellcolor{green!25}136 & \cellcolor{green!25}13.12 & \cellcolor{green!25}\textbf{11.3} \\
\midrule
\multirow{9}{*}{\parbox{3cm}{\centering DiT-XL/2 \\$256 \times 256$\\ (steps = 100 \\ cfg = 4.0)}} & FP32 & 32/32 & 481.65 & 16.75 & 9.82  \\
                                   & SmoothQuant & 4/8 & 181.67 & 17.92 & 25.61 \\
                                   & SmoothQuant & 4/4 & 13.92 & 160.71 & 91.36 \\
                                   & FPQ & 4/8 & \textbf{449.64} & 10.88 & 9.60  \\
                                   & FPQ & 4/4 & 232.92 & 14.69 & 19.78  \\
                                   & GPTQ & 4/8 & 174.26 & 18.24 & 23.41 \\
                                   & GPTQ & 4/4 &  16.99 & 97.37 & 132.07  \\
                                   & \cellcolor{green!25}HQ-DiT & \cellcolor{green!25}4/8 & \cellcolor{green!25}445.12 & \cellcolor{green!25}9.77 & \cellcolor{green!25}\textbf{7.62}  \\
                                   & \cellcolor{green!25}HQ-DiT & \cellcolor{green!25}4/4 & \cellcolor{green!25}437.13 & \cellcolor{green!25}\textbf{9.25} & \cellcolor{green!25}9.94 \\
\midrule
\multirow{4}{*}{\parbox{3cm}{\centering DiT-XL/2\\$512 \times 512$ \\ (steps = 20 \\ cfg = 4.0)}} & FP32 & 4/4 & 430.59 & 16.26 & 10.47 \\
                                   & FPQ & 4/4 & 140.35 & 23.61 & 27.82  \\
                                   & GPTQ & 4/4 & 28.12 & 68.34 & 107.29 \\
                                   & \cellcolor{green!25}HQ-DiT & \cellcolor{green!25}4/4 &\cellcolor{green!25} \textbf{370.69} &\cellcolor{green!25} \textbf{9.44} &\cellcolor{green!25} \textbf{15.91} \\
\midrule
\multirow{1}{*}{\parbox{3cm}{\centering LDM-4\cite{efficientdm}}} & FP32 & 4/4 & 379.19 & 11.71 & 6.08 \\

\bottomrule
\end{tabular}
\end{adjustbox}
\label{tab:condi_results}
\end{table}

We further evaluate the performance of DiT on ImageNet $512 \times 512$. The total number of time steps is set to 20, while the remaining settings remain the same as those for ImageNet $256 \times 256$. According to the evaluation results presented in Table~\ref{tab:condi_results}, our W4A4 model achieves an IS of 370.69, with a 6.82 increase in FID compared to the full-precision model, while obtaining the best performance among the baseline algorithms.
\subsection{Ablation Study}
\label{sec:ablation}
\begin{wrapfigure}{r}{0.6\textwidth}
    \centering
    \includegraphics[width=0.6\textwidth]{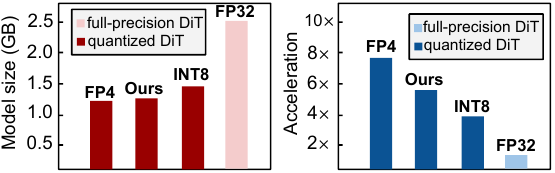}
\caption{Efficiency Analysis of the Quantized Model.}
    \label{figure:efficiency}
\end{wrapfigure}
\paragraph{Impact of Hadamard Transform on Image Quality}
We first investigate the impact of the Hadamard transform on the performance of the quantized model under different settings. Specifically, we evaluate the quality of images generated by HQ-DiT with and without applying the Hadamard transform to the activations. Without Hadamard transform, HQ-DiT fails to generate high-quality images at W4A4 and can just get an IS of 3.14, FID of 200.85. Conversely, by introducing the Hadamard transform, our method achieves an IS of 437.13 and an sFID of 9.94, performance levels comparable to full-precision models.

\paragraph{Impact of FP Composition} Additionally, we conduct experiments to analyze the effect of the FP format selection described in Section~\ref{sec:weight-quant}. Specifically, we compare HQ-DiT with other baselines by fixing the FP composition across the entire DiT layers, resulting three baselines including E3M0, E2M1 and E1M2. The results presented in Table~\ref{tab:fp_composition_results} illustrate that our hybrid FP format achieves an IS of 437.13, outperforming all other FP4 formats, and a FID of 9.25, which is close to the FID result of E1M2 (9.20). However, given the great improvement in the IS score, a degradation of 0.05 in FID is considered insignificant. This demonstrates that our FP composition selection and hybrid FP quantization scheme can effectively enhance the DiT performance.

\begin{table}[ht]
\centering
\caption{The effect of FP format selection. Experiment conducted on ImageNet $256 {\times} \text{ 256}$}
\begin{adjustbox}{width=0.7\columnwidth, center}
\begin{tabular}{@{}cccccccccc@{}} 
\toprule
Model & FP-Format & Bit-width (W/A) & IS $\uparrow$ & FID $\downarrow$\\
\midrule
\multirow{4}{*}{\parbox{3cm}{\centering DiT-XL/2 \\ (steps = 100 \\ cfg = 4.0)}} & E3M0 & 4/4 & 408.88 & 13.32 \\
                                   & E2M1 & 4/4 & 430.33 & 11.65  \\
                                   & E1M2 & 4/4 & 422.77 & \textbf{9.20}  \\
                                   & \cellcolor{green!25}HQ-DiT & \cellcolor{green!25}4/4 & \cellcolor{green!25}\textbf{437.13} & \cellcolor{green!25}9.25 \\

\bottomrule
\end{tabular}
\label{tab:fp_composition_results}
\end{adjustbox}
\end{table}

\subsection{Deployment Efficiency}
Finally, we evaluate the execution cost of the quantized model generated by HQ-DiT by comparing it to the full-precision (FP32) model and the INT quantized model. We evaluate the execution cost in terms of two aspects: model size and theoretical computation time per DiT block. Our computation follows the workflow depicted in Figure~\ref{fig:workflow}. In the evaluation of theoretical computation time, due to the lack of hardware support, we calculated the GFLOPs for each module and estimated the computation time of the quantized model based on the theoretical acceleration performance.
As shown in Figure~\ref{figure:efficiency}, quantization effectively reduces the model size. Although the introduction of online Hadamard transforms increases computational overhead, the overall computational cost is still significantly lower compared to INT8 implementation. More implementation details are provided in the appendix.

\section{Conclusion}
In this paper, we propose HQ-DiT, an efficient data-free PTQ method for low-precision DiT execution. To address the difficulty in quantizing activations, we eliminate outliers by introducing the Hadamard transform with minimal execution cost. To select the appropriate FP format, we propose a selection method based on data statistics. Our experiments demonstrate the superior performance of HQ-DiT compared to other quantization methods. Notably, our 4-bit model achieves higher IS and lower FID compared to the full-precision LDM model.

\newpage 

\bibliographystyle{unsrt}

\begin{thebibliography}{10}

\bibitem{dit}
William Peebles and Saining Xie.
\newblock Scalable diffusion models with transformers.
\newblock In {\em Proceedings of the IEEE/CVF International Conference on Computer Vision}, pages 4195--4205, 2023.

\bibitem{ldm}
Robin Rombach, Andreas Blattmann, Dominik Lorenz, Patrick Esser, and Björn Ommer.
\newblock High-resolution image synthesis with latent diffusion models, 2022.

\bibitem{mo2024dit}
Shentong Mo, Enze Xie, Ruihang Chu, Lanqing Hong, Matthias Niessner, and Zhenguo Li.
\newblock Dit-3d: Exploring plain diffusion transformers for 3d shape generation.
\newblock {\em Advances in Neural Information Processing Systems}, 36, 2024.

\bibitem{feng2024latent}
Shibo Feng, Chunyan Miao, Zhong Zhang, and Peilin Zhao.
\newblock Latent diffusion transformer for probabilistic time series forecasting.
\newblock In {\em Proceedings of the AAAI Conference on Artificial Intelligence}, pages 11979--11987, 2024.

\bibitem{wu2024medsegdiff}
Junde Wu, Wei Ji, Huazhu Fu, Min Xu, Yueming Jin, and Yanwu Xu.
\newblock Medsegdiff-v2: Diffusion-based medical image segmentation with transformer.
\newblock In {\em Proceedings of the AAAI Conference on Artificial Intelligence}, pages 6030--6038, 2024.

\bibitem{gao2023masked}
Shanghua Gao, Pan Zhou, Ming-Ming Cheng, and Shuicheng Yan.
\newblock Masked diffusion transformer is a strong image synthesizer.
\newblock In {\em Proceedings of the IEEE/CVF International Conference on Computer Vision}, pages 23164--23173, 2023.

\bibitem{chen2023pixart}
Junsong Chen, Jincheng Yu, Chongjian Ge, Lewei Yao, Enze Xie, Yue Wu, Zhongdao Wang, James Kwok, Ping Luo, Huchuan Lu, et~al.
\newblock Pixart: Fast training of diffusion transformer for photorealistic text-to-image synthesis.
\newblock {\em arXiv preprint arXiv:2310.00426}, 2023.

\bibitem{Sora}
OpenAI.
\newblock Sora: Creating video from text, 2024.

\bibitem{gao2024lumina}
Peng Gao, Le~Zhuo, Ziyi Lin, Chris Liu, Junsong Chen, Ruoyi Du, Enze Xie, Xu~Luo, Longtian Qiu, Yuhang Zhang, et~al.
\newblock Lumina-t2x: Transforming text into any modality, resolution, and duration via flow-based large diffusion transformers.
\newblock {\em arXiv preprint arXiv:2405.05945}, 2024.

\bibitem{dm_method2}
Prafulla Dhariwal and Alex Nichol.
\newblock Diffusion models beat gans on image synthesis, 2021.

\bibitem{dm_method1}
Fan Bao, Chongxuan Li, Jun Zhu, and Bo~Zhang.
\newblock Analytic-dpm: an analytic estimate of the optimal reverse variance in diffusion probabilistic models, 2022.

\bibitem{nagel2020ptq}
Markus Nagel, Rana~Ali Amjad, Mart Van~Baalen, Christos Louizos, and Tijmen Blankevoort.
\newblock Up or down? adaptive rounding for post-training quantization.
\newblock In {\em Proceedings of the 37th International Conference on Machine Learning}, ICML'20. JMLR.org, 2020.

\bibitem{li2021brecq}
Yuhang Li, Ruihao Gong, Xu~Tan, Yang Yang, Peng Hu, Qi~Zhang, Fengwei Yu, Wei Wang, and Shi Gu.
\newblock {\{}BRECQ{\}}: Pushing the limit of post-training quantization by block reconstruction.
\newblock In {\em International Conference on Learning Representations}, 2021.

\bibitem{lin2021fq}
Yang Lin, Tianyu Zhang, Peiqin Sun, Zheng Li, and Shuchang Zhou.
\newblock Fq-vit: Post-training quantization for fully quantized vision transformer.
\newblock {\em arXiv preprint arXiv:2111.13824}, 2021.

\bibitem{kung2020term}
HT~Kung, Bradley McDanel, and Sai~Qian Zhang.
\newblock Term revealing: Furthering quantization at run time on quantized dnns. corr abs/2007.06389 (2020).
\newblock {\em arXiv preprint arXiv:2007.06389}, 2020.

\bibitem{kung2020term1}
Hsiang-Tsung Kung, Bradley McDanel, and Sai~Qian Zhang.
\newblock Term quantization: Furthering quantization at run time.
\newblock In {\em SC20: International Conference for High Performance Computing, Networking, Storage and Analysis}, pages 1--16. IEEE, 2020.

\bibitem{qdiffusion}
Xiuyu Li, Yijiang Liu, Long Lian, Huanrui Yang, Zhen Dong, Daniel Kang, Shanghang Zhang, and Kurt Keutzer.
\newblock Q-diffusion: Quantizing diffusion models, 2023.

\bibitem{efficientdm}
Yefei He, Jing Liu, Weijia Wu, Hong Zhou, and Bohan Zhuang.
\newblock Efficientdm: Efficient quantization-aware fine-tuning of low-bit diffusion models, 2024.

\bibitem{smoothquant}
Guangxuan Xiao, Ji~Lin, Mickael Seznec, Hao Wu, Julien Demouth, and Song Han.
\newblock Smoothquant: Accurate and efficient post-training quantization for large language models, 2024.

\bibitem{zhang2023afpq}
Yijia Zhang, Sicheng Zhang, Shijie Cao, Dayou Du, Jianyu Wei, Ting Cao, and Ningyi Xu.
\newblock Afpq: Asymmetric floating point quantization for llms.
\newblock {\em arXiv preprint arXiv:2311.01792}, 2023.

\bibitem{frantar2023gptq}
Elias Frantar, Saleh Ashkboos, Torsten Hoefler, and Dan Alistarh.
\newblock Gptq: Accurate post-training quantization for generative pre-trained transformers, 2023.

\bibitem{adaptivefp}
Thierry Tambe, En-Yu Yang, Zishen Wan, Yuntian Deng, Vijay Janapa~Reddi, Alexander Rush, David Brooks, and Gu-Yeon Wei.
\newblock Algorithm-hardware co-design of adaptive floating-point encodings for resilient deep learning inference.
\newblock In {\em 2020 57th ACM/IEEE Design Automation Conference (DAC)}, pages 1--6, 2020.

\bibitem{zhang2022fast}
Sai~Qian Zhang, Bradley McDanel, and HT~Kung.
\newblock Fast: Dnn training under variable precision block floating point with stochastic rounding.
\newblock In {\em 2022 IEEE International Symposium on High-Performance Computer Architecture (HPCA)}, pages 846--860. IEEE, 2022.

\bibitem{H200}
NVIDIA.
\newblock Nvidia blackwell architecture, 2024.

\bibitem{llmfp4}
Shih-yang Liu, Zechun Liu, Xijie Huang, Pingcheng Dong, and Kwang-Ting Cheng.
\newblock Llm-fp4: 4-bit floating-point quantized transformers.
\newblock In {\em Proceedings of the 2023 Conference on Empirical Methods in Natural Language Processing}. Association for Computational Linguistics, 2023.

\bibitem{dettmers2022llmint8}
Tim Dettmers, Mike Lewis, Younes Belkada, and Luke Zettlemoyer.
\newblock Llm.int8(): 8-bit matrix multiplication for transformers at scale, 2022.

\bibitem{ho2020denoising}
Jonathan Ho, Ajay Jain, and Pieter Abbeel.
\newblock Denoising diffusion probabilistic models.
\newblock {\em Advances in neural information processing systems}, 33:6840--6851, 2020.

\bibitem{adm}
Prafulla Dhariwal and Alexander Nichol.
\newblock Diffusion models beat gans on image synthesis.
\newblock {\em Advances in neural information processing systems}, 34:8780--8794, 2021.

\bibitem{ho2022classifier}
Jonathan Ho and Tim Salimans.
\newblock Classifier-free diffusion guidance.
\newblock {\em arXiv preprint arXiv:2207.12598}, 2022.

\bibitem{sohldickstein2015deep}
Jascha Sohl-Dickstein, Eric~A. Weiss, Niru Maheswaranathan, and Surya Ganguli.
\newblock Deep unsupervised learning using nonequilibrium thermodynamics, 2015.

\bibitem{nichol2021glide}
Alex Nichol, Prafulla Dhariwal, Aditya Ramesh, Pranav Shyam, Pamela Mishkin, Bob McGrew, Ilya Sutskever, and Mark Chen.
\newblock Glide: Towards photorealistic image generation and editing with text-guided diffusion models.
\newblock {\em arXiv preprint arXiv:2112.10741}, 2021.

\bibitem{ramesh2022hierarchical}
Aditya Ramesh, Prafulla Dhariwal, Alex Nichol, Casey Chu, and Mark Chen.
\newblock Hierarchical text-conditional image generation with clip latents, 2022.

\bibitem{ptqd}
Yefei He, Luping Liu, Jing Liu, Weijia Wu, Hong Zhou, and Bohan Zhuang.
\newblock Ptqd: Accurate post-training quantization for diffusion models, 2023.

\bibitem{xu2023qa}
Yuhui Xu, Lingxi Xie, Xiaotao Gu, Xin Chen, Heng Chang, Hengheng Zhang, Zhensu Chen, Xiaopeng Zhang, and Qi~Tian.
\newblock Qa-lora: Quantization-aware low-rank adaptation of large language models.
\newblock {\em arXiv preprint arXiv:2309.14717}, 2023.

\bibitem{han2024parameter}
Zeyu Han, Chao Gao, Jinyang Liu, Sai~Qian Zhang, et~al.
\newblock Parameter-efficient fine-tuning for large models: A comprehensive survey.
\newblock {\em arXiv preprint arXiv:2403.14608}, 2024.

\bibitem{bfloat}
Bfloat16: The secret to high performance on cloud tpus.
\newblock \url{https://cloud.google.com/blog/products/ai-machine-learning/bfloat16-the-secret-to-high-performance-on-cloud-tpus}.
\newblock Accessed: 2021-03-29.

\bibitem{tf32}
Accelerating ai training with nvidia tf32 tensor cores.
\newblock \url{https://developer.nvidia.com/blog/accelerating-ai-training-with-tf32-tensor-cores/}.
\newblock Accessed: 2021-03-29.

\bibitem{quip}
Albert Tseng, Jerry Chee, Qingyao Sun, Volodymyr Kuleshov, and Christopher~De Sa.
\newblock Quip\#: Even better llm quantization with hadamard incoherence and lattice codebooks, 2024.

\bibitem{DiT1}
Facebook Research.
\newblock Scalable diffusion models with transformers (dit), 2022.

\bibitem{ho2022video}
Jonathan Ho, Tim Salimans, Alexey Gritsenko, William Chan, Mohammad Norouzi, and David~J Fleet.
\newblock Video diffusion models.
\newblock {\em Advances in Neural Information Processing Systems}, 35:8633--8646, 2022.

\bibitem{zhang2023adding}
Lvmin Zhang, Anyi Rao, and Maneesh Agrawala.
\newblock Adding conditional control to text-to-image diffusion models.
\newblock In {\em Proceedings of the IEEE/CVF International Conference on Computer Vision}, pages 3836--3847, 2023.

\end{thebibliography}






\newpage
\appendix

\section{Implement Details}
In our experiment, we adopt the E2M1 format for quantizating activation. For the quantization of weight matrices, we compare different floating-point formats with our proposed floating-point format selection method. We use a block size of 64 during the quantization process with a calibration dataset of 512. All experiments are conducted on Nvidia A100 GPUs. Furthermore, the rest baseline algorithms, including SmoothQuant, GPTQ, and FPQ, we adopt the same calibration dataset as HQ-DiT. All the rest settings, including time information at each timestep, category labels, random noise, and classifier-free guidance are also kept the same.

In the implementation of SmoothQuant, we collect the maximum values of the activation and weight matrices from the first layers of the attention module and the FFN module in the forward propagation backbone network (specifically, the query, key, and value calculations in the attention module, and fc1 in the FFN). We set the hyperparameter $\alpha = 0.5$ to determine the correction factor and conduct related experiments based on this setup.

For the GPTQ method, we perform experiments on the ImageNet dataset with resolutions of $256 \times 256$ and $512 \times 512$. In the $256 \times 256$ experiments, we select a block size of 64, while for the $512 \times 512$ experiments, we use a block size of 16. We quantize and update the weights of the matrices in each linear layer of the forward propagation backbone network according to the GPTQ method. Though GPTQ method is originally designed for Linear quantization, we find it suitable for float point quantization. The experimental results, focusing solely on quantizing weight matrices using the GPTQ method, are provided in Table \ref{tab:gptq}.

To implement the FPQ method, we consider all floating-point formats except those with an exponent bit of 0 as the search range, which is equivalent to the linear quantization. For each linear layer, we search for the optimal floating-point format and the optimal pre-shifted bias according to the method proposed in the FPQ paper.

To generate the evaluation results, we adopt the same sampling process across all the baseline algorithms. We randomly sampled 50K images and calculated evaluation metrics involving IS, FID, and sFID using the evaluation suite.

\begin{table}[H]
\centering
\caption{GPTQ results for conditional image generation. Experiment conducted on ImageNet}
\begin{adjustbox}{width=0.7\columnwidth, center}
\begin{tabular}{@{}cccccc@{}} 
\toprule
Model & Method & Bit-width (W/A) & IS \(\uparrow\) & FID \(\downarrow\) &sFID \(\downarrow\) \\
\midrule
\multirow{2}{*}{\parbox{3cm}{\centering $256 \times 256$\\cfg=1.5}}
& FP32 & 32/32 & 266.57 & 2.55 & 5.34
\\& GPTQ & 4/32 & 210.8 & 5.20 & 8.98  \\
\midrule
\multirow{2}{*}{\parbox{3cm}{\centering $256 \times 256$\\cfg=4.0}}
& FP32 & 32/32 & 481.65 & 16.75 & 9.82
\\& GPTQ & 4/32 & 447.12 & 11.70 & 7.83  \\
\midrule
\multirow{2}{*}{\parbox{3cm}{\centering $512 \times 512$\\cfg=4.0}}
& FP32 & 32/32 & 430.59 & 16.26 & 10.47
\\& GPTQ & 4/32 & 384.46 & 11.64 & 14.38  \\

\bottomrule
\end{tabular}
\label{tab:gptq}
\end{adjustbox}
\end{table}

\section{Activation Distribution in DiT}
\begin{figure}[htbp]
\centering
\includegraphics[scale=0.5]{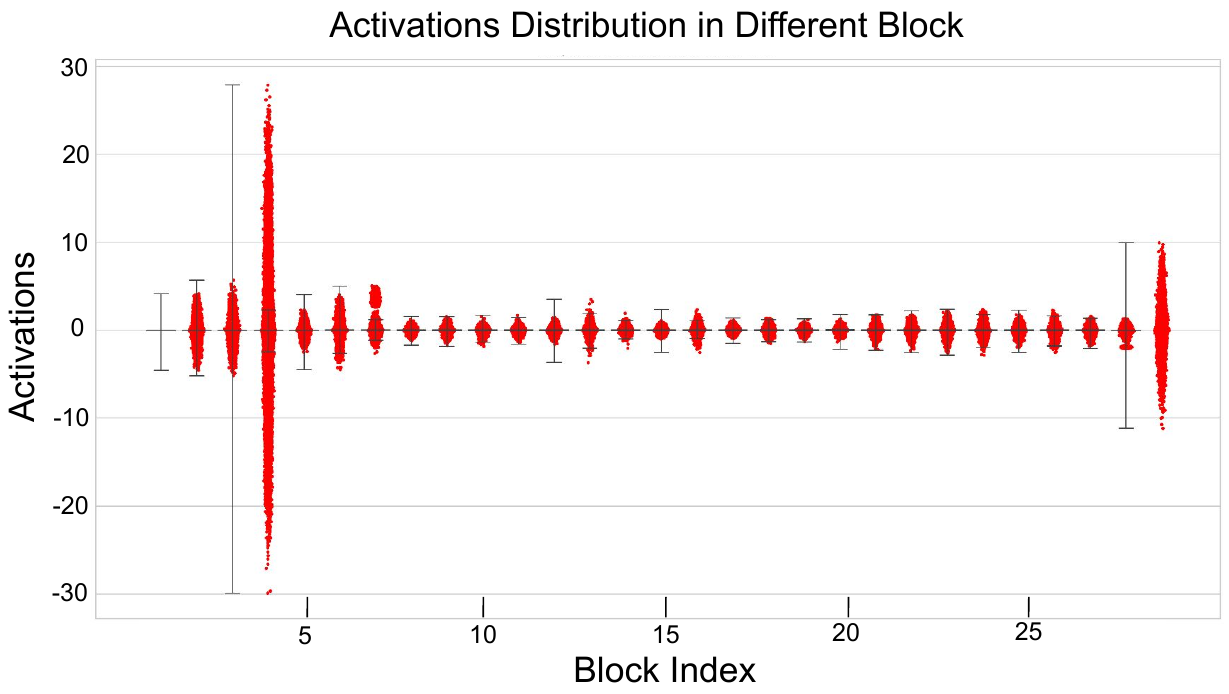}
\caption{Magnitude of the activations in different blocks}
\label{figure:blocks}
\end{figure}

In this section, we present the range on of DiT activations of across different blocks on the ImageNet $256 \times 256$. The distribution of activation can be found Figure~\ref{figure:blocks}. It is noticeable that the original DiTs' activations exhibit high variance between different blocks. By applying the Hadamard transform, outliers in the activations can be effectively eliminated, leading to an accurate per-channel activation quantization.

\section{Fast Hadamard Transform}
A Hadamard matrix is an orthogonal matrix whose entries are proportional to \(\{+1, -1\}\). For matrices of dimension \(2^n\), a Hadamard matrix can be constructed. Specifically, the matrix \(H_{2^n}\) is constructed using the Kronecker product:
\[
H_2 = 
\begin{bmatrix}
1 & 1 \\
1 & -1 
\end{bmatrix}
,
and\ H_{2^n}=
\begin{bmatrix}
H_{2^{n-1}} & H_{2^{n-1}} \\
H_{2^{n-1}} & -H_{2^{n-1}} 
\end{bmatrix}
\]

For any matrix \(X \in \mathbb{R}^{m \times n}\) and orthogonal Hadamard matrix \(H \in \mathbb{R}^{n \times n}\), their matrix multiplication will cause a computing complexity of \(\Theta(mn^2)\). When \( n \) is a power of 2, we can apply the fast Walsh-Hadamard transform to compute \( XH \). Utilizing a divide-and-conquer strategy, we recursively decompose the Hadamard matrix of size \( n \) into two matrices of size \( \frac{n}{2} \). This allows the computation to be completed with \( O(mn \log n) \) additions and subtractions.

For dimension \(n\) that is not powers of 2, we can factorize \(n\) as \(n=pq\), where \(p\) is the largest power of 2 such that there exists a known Hadamard matrix of size \(p\). Then, the Hadamard matrix \(H\) of order \(n\) can be constructed as:
\[
H_{n} = H_p \otimes H_q
\]
where \(H_p\) and \(H_q\) are Hadamard matrices of order \(p\) and \(q\), respectively. Then, we can compute \( XH \) using a divide-and-conquer strategy similar to the fast Walsh-Hadamard transform. This involves recursively processing \( H_p \), allowing the computation to be executed with \( O(mnq \log p) \) additions and subtractions. For instance, for a dimension of \( 28672 = 1024 \times 28 \), this method can significantly reduce the computational overhead.

\newpage
\section{Additional Sample Visualization}
In this section, we provide random samples from W4A4 quantized model based on FPQ and our method HQ-DiT. Results are shown in figures below. 
\begin{figure}[htbp]
\centering 
\subfigure[FPQ]{
\begin{minipage}{12cm}
\centering    
\includegraphics[width=12cm, height=9cm]{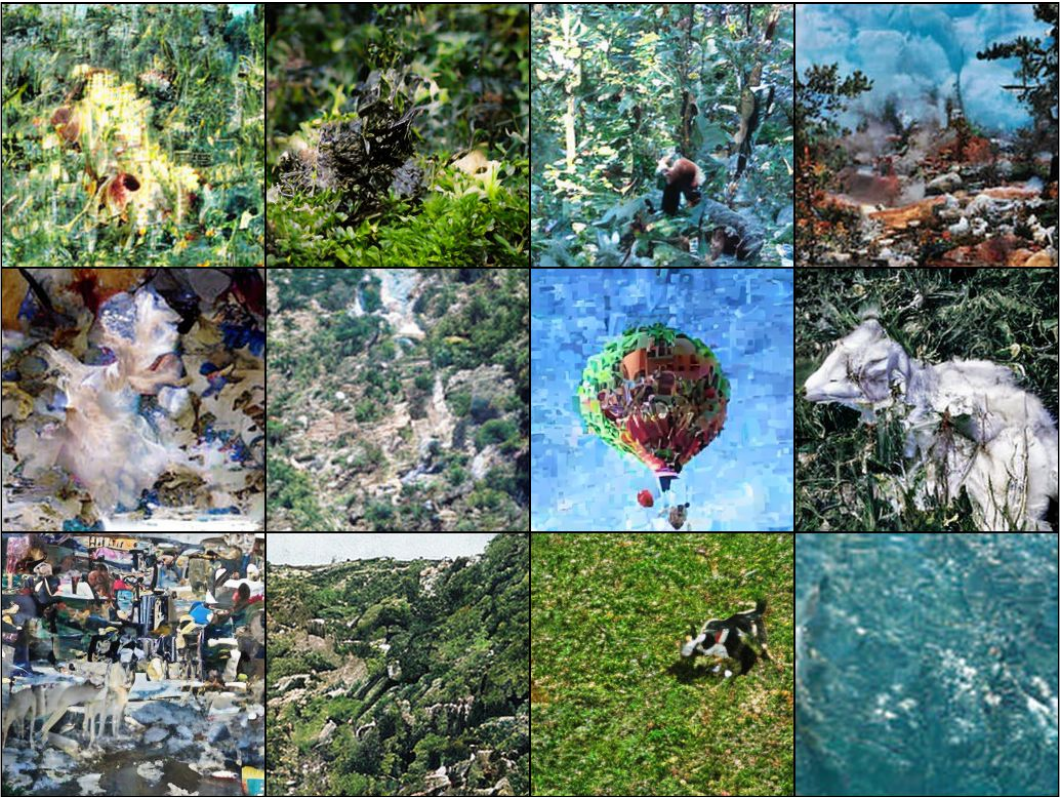} 
\end{minipage}
}
\subfigure[Ours]{ 
\begin{minipage}{12cm}
\centering   
\includegraphics[width=12cm, height=9cm]{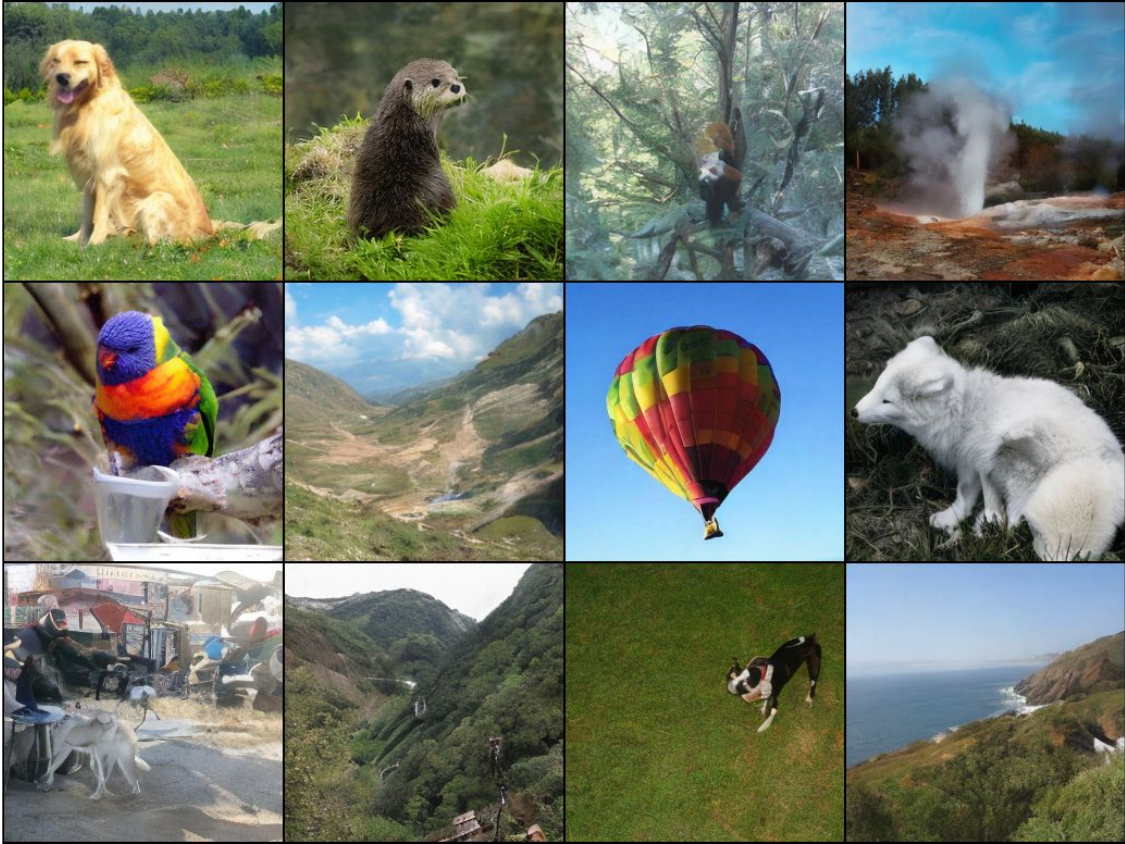} 
\end{minipage}
}
\caption{Samples are generated by W4A4 DiT model on ImageNet $256 \times 256$, cfg is set to 1.5.}    
\label{sample:1}  
\end{figure}

\begin{figure}[htbp]
\centering 
\subfigure[FPQ]{
\begin{minipage}{12cm}
\centering    
\includegraphics[width=12cm, height=9cm]{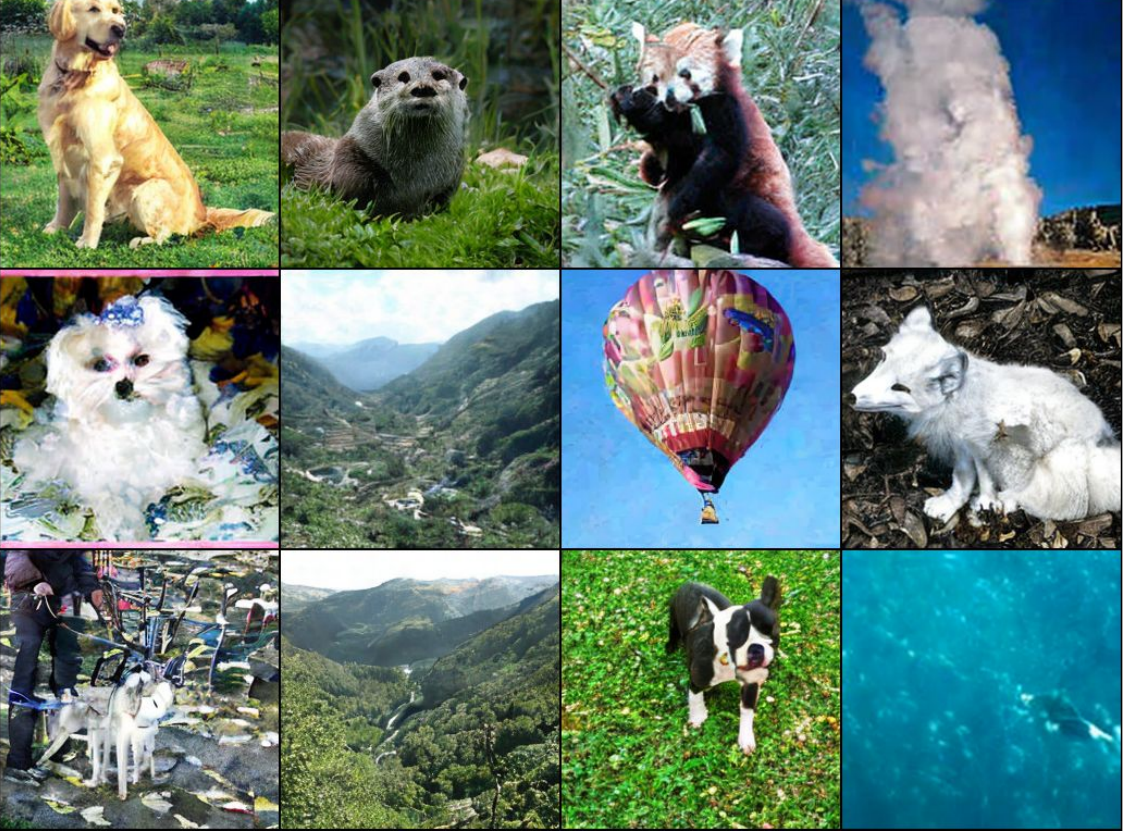} 
\end{minipage}
}
\subfigure[Ours]{ 
\begin{minipage}{12cm}
\centering   
\includegraphics[width=12cm, height=9cm]{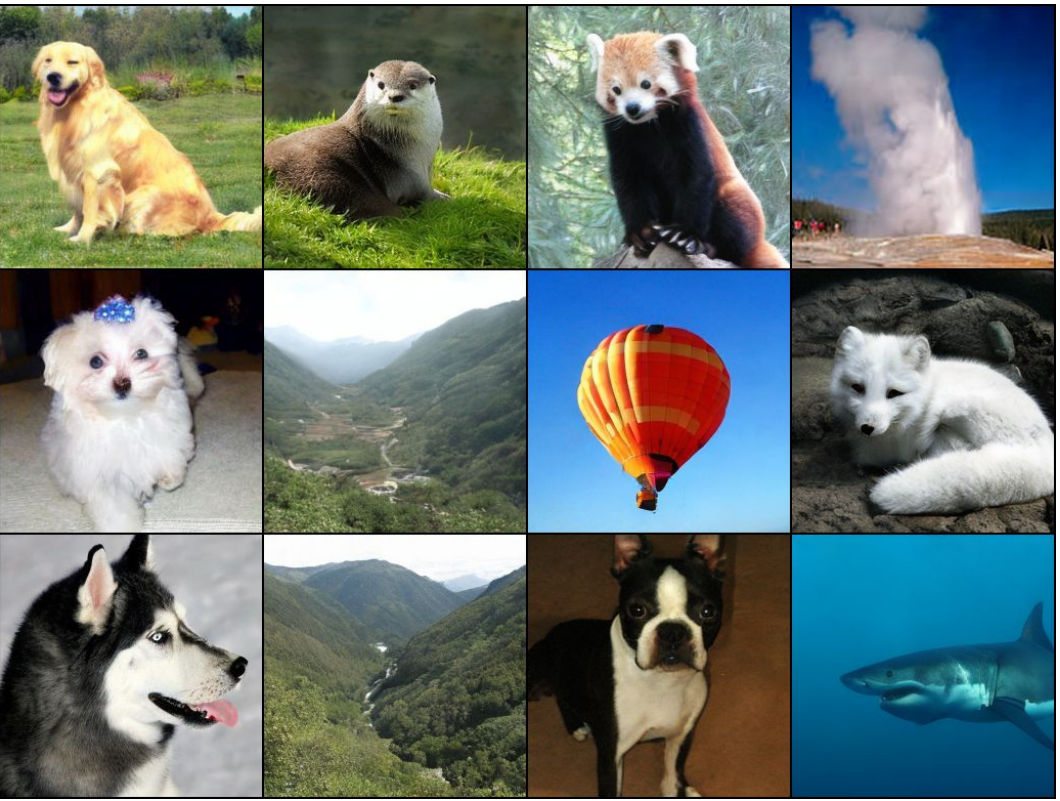} 
\end{minipage}
}
\caption{Samples generated by W4A4 DiT model on ImageNet $256 \times 256$ (cfg=4.0)}    
\label{sample:2}    
\end{figure}

\begin{figure}[htbp]
\centering 
\subfigure[FPQ]{
\begin{minipage}{12cm}
\centering    
\includegraphics[width=12cm, height=9cm]{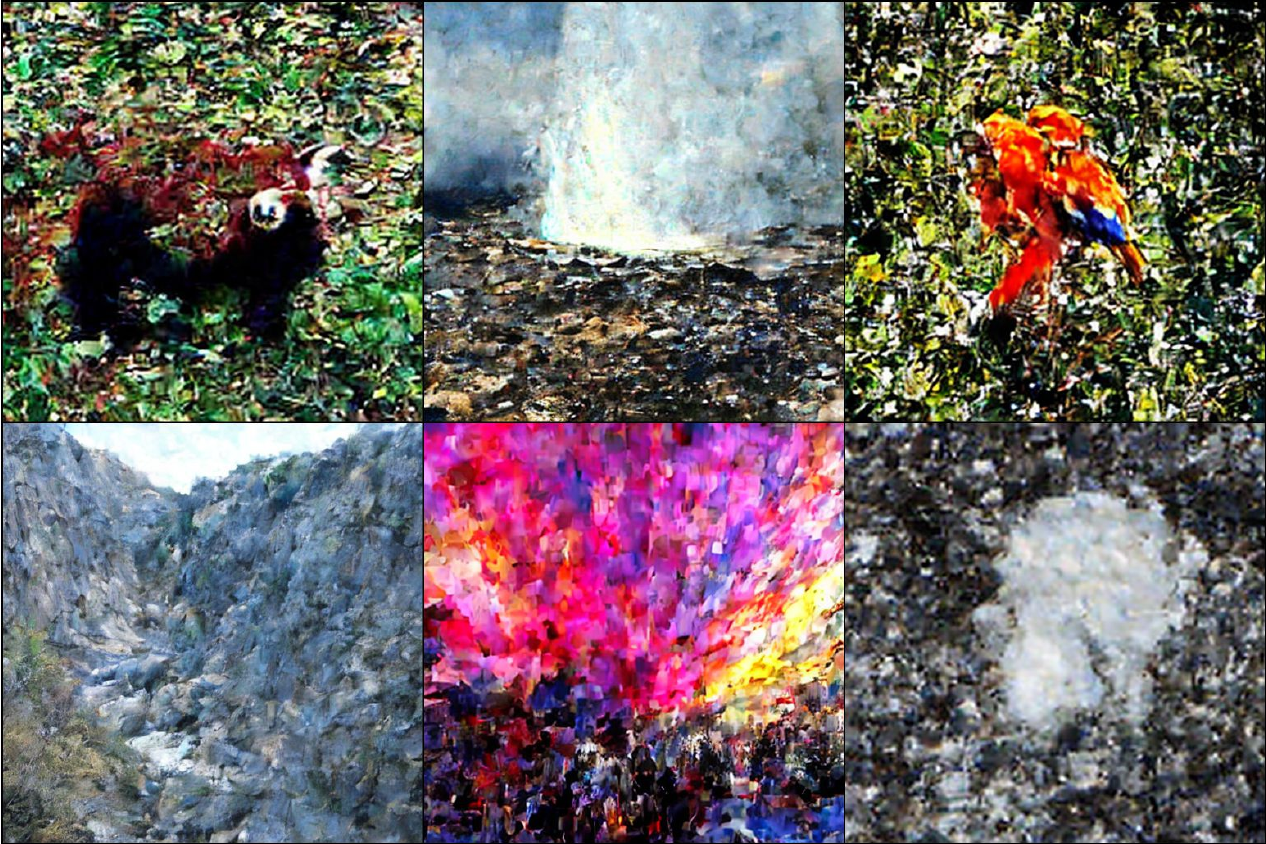} 
\end{minipage}
}
\subfigure[Ours]{ 
\begin{minipage}{12cm}
\centering   
\includegraphics[width=12cm, height=9cm]{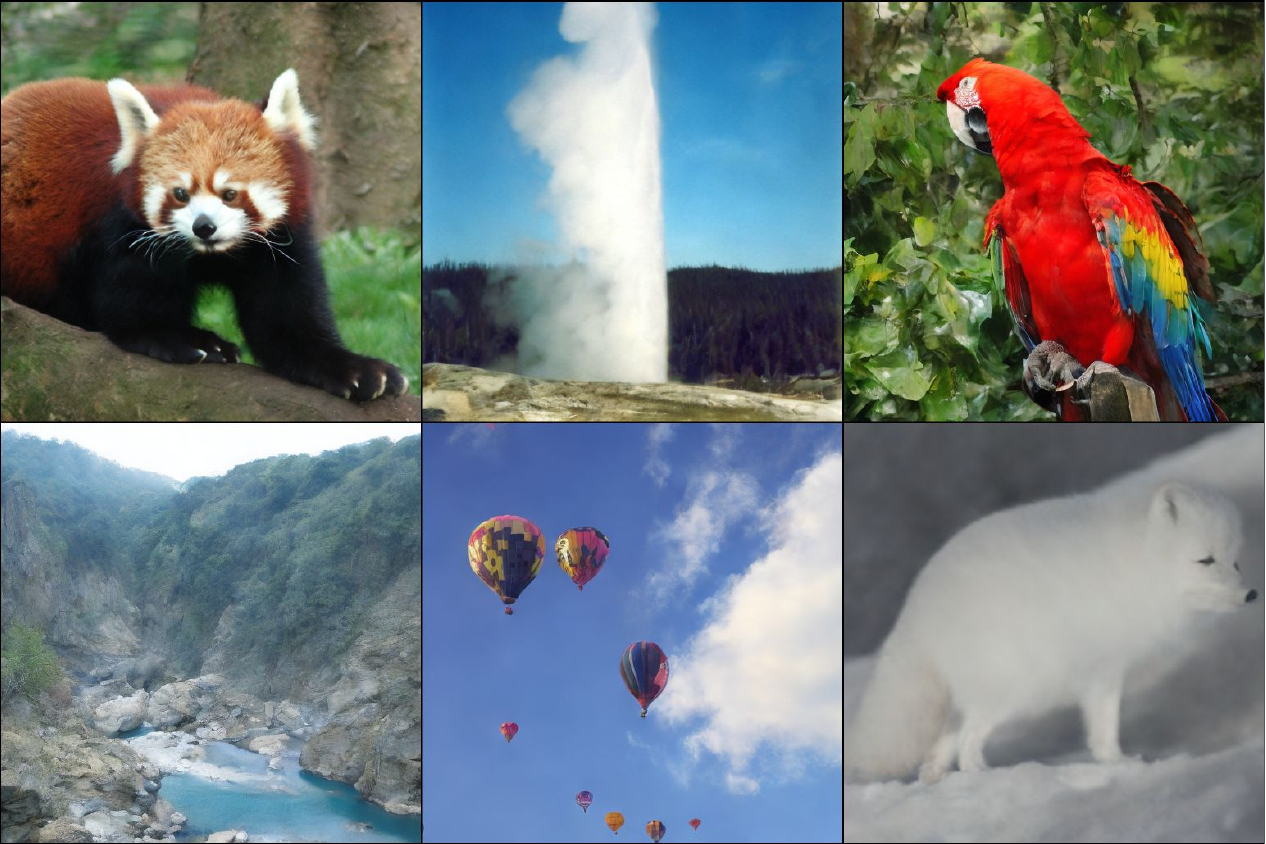} 
\end{minipage}
}
\caption{Samples generated by W4A4 DiT model on ImageNet $512 \times 512$ (cfg=4.0)}    
\label{sample:3}    
\end{figure}

\end{document}